\documentclass{article}
\usepackage{arxiv}
\usepackage{upgreek}
\usepackage[utf8]{inputenc} 
\usepackage[T1]{fontenc}    
\usepackage{hyperref}       
\usepackage{url}            
\usepackage{booktabs}       
\usepackage{amsfonts}       
\usepackage{nicefrac}       
\usepackage{microtype}      
\usepackage{lipsum}		
\usepackage{graphicx}
\usepackage{doi}

\title{Deep learning for prediction of hepatocellular carcinoma recurrence after resection or liver transplantation: a discovery and validation study}

\author{ {Zhikun Liu\footnotemark[1]}\\
	Department of Hepatobiliary and Pancreatic Surgery\\
	The Center for Integrated Oncology and Precision Medicine\\
	Affiliated Hangzhou First People’s Hospital, Zhejiang University School of Medicine\\
	Hangzhou, 310006, China \\
	\And
	{Yuanpeng Liu\footnotemark[1]} \\
	Department of Electrical Engineering and Computer Science\\
	Syracuse University\\
	Syracuse, NY, 13244-4100, USA \\
	\And
	{Yuan Hong}\\
	School of Mathematical Sciences\\
	 Zhejiang University\\
	  Hangzhou, 310058, China\\
	\And
	{Jinwen Meng}\\
	Department of Hepatobiliary and Pancreatic Surgery\\
	The Center for Integrated Oncology and Precision Medicine\\
	 Affiliated Hangzhou First People’s Hospital, Zhejiang University School of Medicine\\
	Hangzhou, 310006, China\\  
	\And
	{Jianguo Wang}\\
	Department of Hepatobiliary and Pancreatic Surgery\\
	The Center for Integrated Oncology and Precision Medicine\\
	Affiliated Hangzhou First People’s Hospital, Zhejiang University School of Medicine\\
	Hangzhou, 310006, China\\  
	\And
	{Shusen Zheng}\\
	Department of Hepatobiliary and Pancreatic Surgery\\
	The First Affiliated Hospital, Zhejiang University School of Medicine\\
	Hangzhou, 310006, China\\  
	NHC Key Laboratory of Combined Multi-organ Transplantation\\
	Hangzhou, 310003, China\\
	\And
	{Xiao Xu\footnotemark[2]}\\ 
	Department of Hepatobiliary and Pancreatic Surgery\\
	The Center for Integrated Oncology and Precision Medicine\\
	Affiliated Hangzhou First People’s Hospital, Zhejiang University School of Medicine\\
	Hangzhou, 310006, China\\  
	NHC Key Laboratory of Combined Multi-organ Transplantation\\
	Hangzhou, 310003, China\\
}

\renewcommand{\shorttitle}{\textit{arXiv} Template}

\hypersetup{
pdftitle={A template for the arxiv style},
pdfsubject={q-bio.NC, q-bio.QM},
pdfauthor={David S.~Hippocampus, Elias D.~Striatum},
pdfkeywords={First keyword, Second keyword, More},
}
\begin{document}
\maketitle
\footnotetext[1]{Zhikun Liu and Yuanpeng Liu contributed equally to this work.}
\footnotetext[2]{Corresponding authors: zjxu@zju.edu.cn}

\begin{abstract}
	Background and aim: Improved prognosis classifiers are needed to stratify patients with hepatocellular carcinoma (HCC). This study aimed to develop a classifier of prognosis after resection or liver transplantation (LT) for HCC using deep learning on the ubiquitously available histological images.\\
	
	Methods: A deep learning model was developed on 1118 patients across four independent cohorts. The Nucleus map set (n=120) was used to train U-net to capture the nuclear architecture. The Train set (n=552) included HCC patients treated by resection and had a distinct outcome. The LT set (n=144) contained patients with HCC treated by LT. The Train set and its nuclear architectural information extracted by U-net were used to train the MobileNet V2 based classifier (MobileNetV2\_HCC\_Class), purpose-built for classifying supersized heterogeneous images. The classifier was independently tested on the LT set, and then externally validated on the TCGA set (n=302). The primary outcome was recurrence- free survival (RFS).\\
	
	Results: The MobileNetV2\_HCC\_Class was a strong predictor of RFS in both LT set and TCGA set, even after stratification for other common prognostic features. The classifier provided a hazard ratio of 3.44 (95\% CI 2.01-5.87, p<0.001) for high risk versus low risk in LT set, and 2.55 (95\% CI: 1.64-3.99, p<0.001) after adjusting for established prognostic factors significant in univariable analyses of the same cohort. The MobileNetV2\_HCC\_Class maintained relatively higher discriminatory power (time-dependent AUC) than the other factors after LT or resection in the independent validation set. Pathological review showed that the tumoral areas most predictive of recurrence were characterized by the presence of stroma, high degree of cytological atypia, nuclear hyperchromasia, and a lack of immune cell infiltration.\\
	
	Conclusion: A clinically useful prognostic classifier was developed using deep learning allied to histological slides. The classifier assists in refining the prognostic prediction of HCC patients and identifying patients who would benefit from more intensive management.
\end{abstract}


\section{Introduction}
\lipsum[2]
\lipsum[3]

\section{Introduction}
\label{sec:headings}
Hepatocellular carcinoma (HCC) is the seventh most common solid malignancy and the third leading cause for cancer-related deaths worldwide\cite{bray2018global}. The prevalence of HCC is relatively high in the Asia-Pacific countries\cite{omata2017asia}. Hepatectomy and liver transplantation (LT) remains the main therapy for HCC. Despite significant progress in the diagnostic and management techniques of HCC, the recurrence rates remain as high as 70-80\% after hepatectomy and 20-40\% after LT \cite{kudo2017systemic}\cite{Llovet}\cite{llovet2019randomized}\cite{kulik2018therapies}. Refinement of prognostic models, especially those based on the accessible data, could easily allow attending to the early warning signs during follow up and prolonging adjuvant therapeutic decisions\cite{fujiwara2018risk}.

Prognosis is strongly related to pathological features. The histological analysis of tumor tissues certainly provides crucial information for the stratification of patients and treatment allocation. Histological slides contain a vast amount of information that can be quantitatively assessed by deep learning algorithms. Convolutional neural networks (CNNs) are widely employed in the fields of speech recognition, traffic sign management, and face recognition\cite{sainath2015deep}\cite{shao2018real}\cite{guo2020multi}. CNNs have passed in many image interpretation tasks and could retrieve additional information from histopathology images. In a recent pioneering study, simulating routine pathology workflows, a subset of deep learning-based algorithms, achieved better diagnostic performance than a panel of 11 pathologists at detecting lymph node metastases of breast cancers\cite{bejnordi2017diagnostic}. Coudray et al showed that CNNs can diagnose the main histological subtypes of non-small cell lung cancer and predict the mutational status of genes (e.g. STK11, EGFR)\cite{coudray2018classification}. CNNs have also been reported to predict the aggressiveness of colorectal cancer\cite{kather2019predicting}. More recently, growing evidence suggests that the computational processing of histological slides will refine the prediction of patient prognosis, thereby improving treatment allocation. A deep learning based model by Saillard et al could predict the survival in HCC patient. In their study the features were extracted from the images by pre-trained CNNs and the network then selected 25 tiles with the highest and lowest scores for the prediction of patient survival\cite{saillard2020predicting}. Skrede et al successfully developed a marker to predict the prognosis of colorectal cancer in the large cohorts using MobileNet V2, one of CNNs\cite{skrede2020deep}, building the model by Multiple Instance Learning (MIL). Additionally, the local spatial arrangement of nuclei in histopathology images has been shown to have prognostic value in oropharyngeal cancers\cite{lu2021feature}.

In this study, four independent cohorts of post-operation HCC patients were investigated to develop and validate MobileNet V2 based model for the improved prediction of prognosis in HCC. The scientific or innovation points of our method comes from two aspects: like Skrede, trained MobileNet V2 by MIL which allow training on tile collections labelled with the label of its whole-slide image\cite{skrede2020deep}, and use of nuclear architectural information in building model, which was useful in cancer grading and predicting patient outcomes\cite{ji2019nuclear}. In order to capture localized nuclear architectural information, local nuclei measurements were constructed by U-net in the independent cohort. Here we show that the models predict survival more accurately than the clinical and pathological features. The aim of this study was to use MobileNet V2 to analyze pathology images and to develop an automatic prognostic classifier for HCC patients treated with liver resection. In addition, we generalized the prognostic power of the MobileNet V2 via validation across different cohorts, even after LT.

\section{Methods}
\subsection{Patients and samples}

Four different cohorts were used in this study. The stained tumor tissue sections with adequate quality and titles were used. The first cohort was used to train U-net to capture the localized nuclear architectural information (Nucleus map set, n=120). The second cohort was from HCC patients treated by surgical resection at the First hospital of Zhejiang University between 2010 and 2016 and have a so-called distinct outcome (Train set, n=552). To obtain clear ground truth, patients with the distinct outcome, either good or poor (good: 274, poor: 278), were used as training cohorts. Patients with a 4 years follow-up after resection and no record of recurrence were assigned to the good outcome group. While the poor outcome group consisted of patients relapsing within 1.6 years (exclusive) after surgery. The third cohort from patients with HCC underwent liver transplantation at the First hospital of Zhejiang University between 2015 and 2019 (LT set, n=144). Nucleus map set, Train set, and LT set were collated from three different batches of HCC patients after obtaining the respective approval from the ethics committee of the institution. The fourth dataset, namely TCGA set with complete follow-up data (n=302), was included for external validation.
Nucleus map set, Train set, and LT set were scanned and digitized using a P250FLASH2 (3DHISTECH3) at 20x magnification. Nucleus map set was employed for training U-net. Train set was used to train MobileNet V2. LT set was used to externally validate the model in HCC treated by LT. The histology slides, clinical follow-up data and histology annotation were retrieved from TCGA database \footnote{https://cancergenome.nih.gov/}.

\subsubsection{Titles cropped and color normalization}
Due to the limitations of graphic card memory, it is almost impossible to process the whole-sliced pathological images, which are usually in a resolution of 100000 by 100000, on GPU or main memory during training phase. Current best practice is to cut the large image into hundreds of small pieces, which are called tiles or patches\cite{skrede2020deep}\cite{corredor2019spatial}\cite{kather2019deep}. The titles were 512 $times$ 512 pixels (px) and 0.25 $\upmu$m per px, cropped from Nucleus map set, Train set, and LT set, and resized to a resolution of 224 × 224. The titles were normalized as described previously\cite{tam2016method}.

\subsection{Extending features with segmentation heat map of nuclear architectural information by U-net}
Before feeding data into the model, we used a trained image segmentation model to get the heat map of nuclei segmentation for each tile. The segmentation model is a U-net neural network trained with Nucleus map set. Let I denote an image slice, p is the output of the U-net, y is the ground-truth label of the image slice and $\epsilon$=0.00000001 is a smoothing term to make the denominator non-zero. Loss function is Dice loss\ref{dice} and final Dice Score on the TCGA test set can reach 82\%. The segmentation result is not desired to be too perfect, since information other than nuclei, such as cytoplasm and shape of the whole cell, is also ponderable in the heat map.

\begin{equation}
	\label{dice}
	L_{dice} = 1-2\times \frac{\sum_{i \in I}{p_i y_i}+\epsilon}{\sum_{i \in I}{p_i} + \sum_{i \in I}{y_i} + \epsilon}
\end{equation}

\subsection{Realization of MIL in MobileNet V2}
The main guiding methodology of our work is MIL, which is a kind of weak supervised learning to deal with lack of annotations. All the tiles can be fed to train the learning model. However, such approach has a serious drawback in classification work. In many cases, the content of one small tile conflicts with the label of the original pathological image, especially in HCC with great heterogeneity. To solve this problem, MobileNet V2 was developed using MIL to allow training on tile collections labeled with the label of its whole-slide image. We used MIL to take advantage of features from every tile. Instead of annotating every tile with their ancestor’s label and dumping them into to network directly, we packed all the tiles into a bag that has the same label as the original pathological image’s. A bag, which represents a pathological image, passes through a trained neural network to calculate scores of each tile in the bag, and an aggregational function produces a weight-average score for the bag. By setting a threshold, the pathological image can be classified into a certain class.

Each 224$\times$224 tile was color-normalized in Vahadane method\cite{vahadane2016structure}. After nuclei segmentation, the color-normalized RGB tiles were concatenated with its heat map in channel level and produced a 4-channel tile. Then, bags of 4-channel tiles were dumped into a feature extractor, which is a MobileNet V2 model, and score of each tile was calculated. Generalized mean was used as the aggregation function since it is able to keep the extremes while taking account the average.

The aggregation function reads as\ref{agg}, where p is a hyper parameter.
\begin{equation}
	\label{agg}
	S = (\sum {s_{i}^{p}})^{\frac{1}{p}}
\end{equation}

Output of the aggregation function, which represents score of the pathological image, was activated by a sigmoid function and compared with a given threshold t, where t is also a hyper parameter, and finally be classified into a certain class.

\subsection{Training strategy}
During the training process, we deployed a decay learning rate, which is initiated with 0.0001 and halves every 10 epochs. Due to the limitation of GPU memory, training batch size could only be set to 1. Besides, threshold t is 0.4457 and p in aggregation function is 3. Cross-Entropy with L2 regulation \ref{p}--\ref{ce} is selected as the loss function and regularized factor $\alpha$ is 0.02.

\begin{equation}
	\label{p}
	p_i=\left\{
	\begin{array}{rcl}
		(\frac{t-s_i}{t}+1)\cdot 0.5 & & {s_i \leq t}\\
		(1-\frac{s_i-t}{1-t}) \cdot 0.5 & & {s_i \textgreater t}
	\end{array} \right.
\end{equation}

\begin{equation}
	\label{ce}
	L(\{p_i\}, \{y_i\}, W) = -\frac{1}{N}(y_i \cdot \log(p_i) + (1-y_i) \cdot \log(1-p_i)) + \alpha \sum_{w_i \in W} w^2
\end{equation}

\subsection{Analysis of tiles with high predictive value}
In order to provide insights into the features associated with tumor aggressiveness, tiles with high and low risk scores were extracted and further analyzed. In total, 4 histological features of tumoral liver tissues were systematically recorded.

\subsection{Statistical analysis}
Log-rank tests were used to compare the survival distributions between stratification subgroups. We used time-dependent AUC (area under ROC curve) as a metric for assessing the predictive performance of our model and baseline clinical, biological and pathological features. Statistical analyses were performed in R (version 3.6.0) using ggplot2, survival, and survminer packages. The training and deployment of CNNs were conducted with Python using a standard desktop workstation (Nvidia Tesla P40 GPUs each with 24GM memory). P-value $\textless$ 0.05 was deemed as statistically significant.

\section{Reseult}
\label{sec:others}

\subsection{Patient Characteristics and Model development}
The Nucleus map set was used to train U-net to capture the localized nuclear architectural information (n=120). The other three sets were used for training and validation: 552 patients from the Train set were used to develop the model, 144 patients from the LT set, and the TCGA set was used to externally validate the model. 552 patients from the Train set had a distinct outcome (good: 274, poor: 278) and were used as a training cohort to obtain clear ground truth. Patient demographics are summarized in table 1.

\begin{figure*}[h]
	\centerline{\includegraphics[width=5 in]{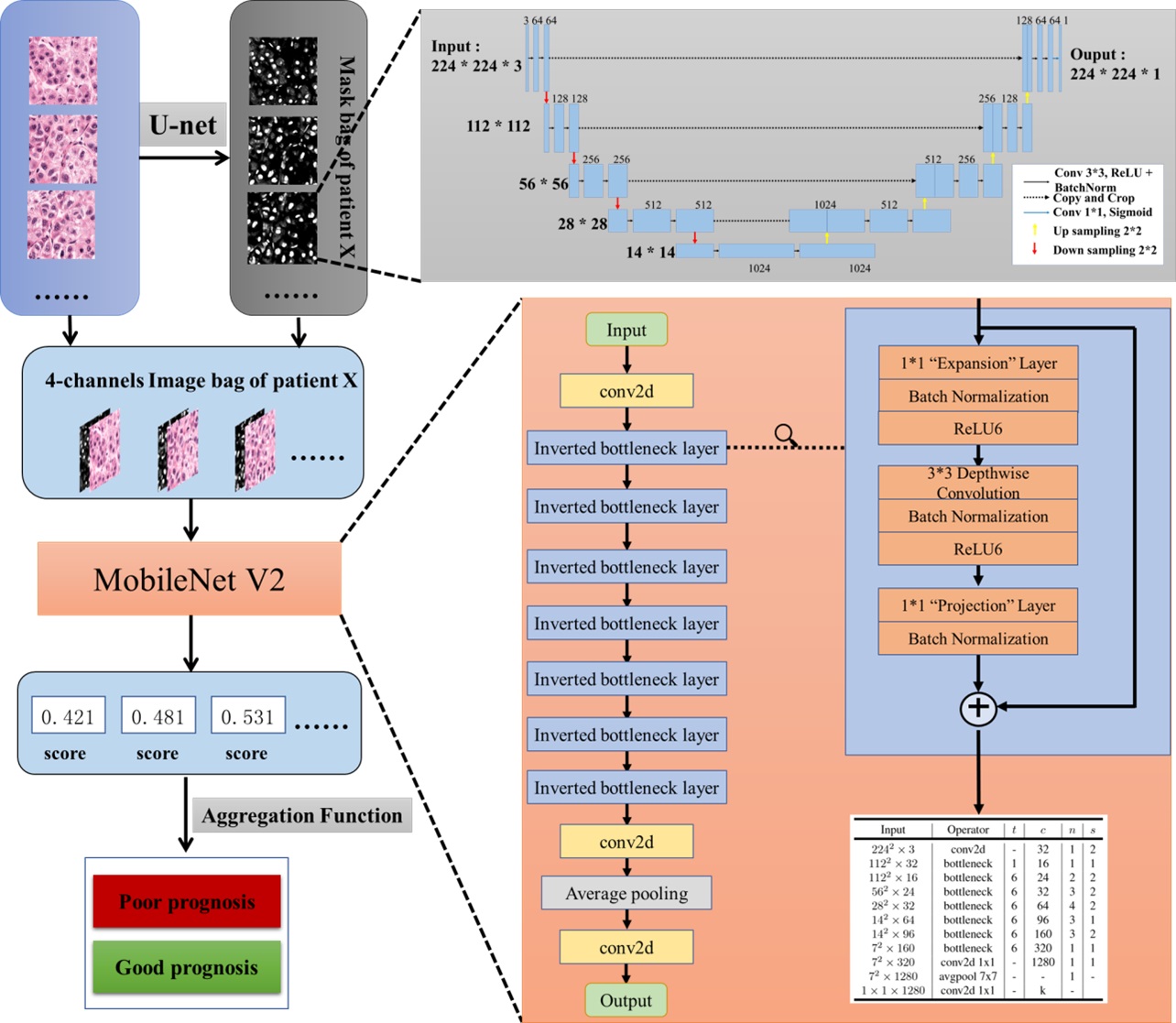}}
	\caption{Pipeline of MobileNetV2\_HCC\_Class. Small image patches of 224$\times$224 pixels from Train set and the heat map of nuclei segmentation for each tile by a pretrained U-net. Concatenate the heatmap of nuclei segmentation and the color-normalized RGB tiles in channel level and produce a 4-channel tile. Then bags of 4-channel tiles are dump into a feature extractor of the MobileNetV2 model. We use generalized mean with sign as the aggregation function since it is able to keep the extremes while taking account the average, and finally be classified into certain class.
	}
	\label{pip}
\end{figure*}

Firstly, we used an image segmentation model to get the heatmap of nuclei segmentation for each tile. The segmentation model is a U-net neural network trained with Nucleus map set. Loss function is Dice and final Dice Score on Nucleus map set can reach 82\%. The segmentation result is not desired to be too precise, since information other than nuclei, such as cytoplasm and shape of the whole cell, is also ponderable in the heatmap. A total of 57415 tiles (small image patches of 224$\times $224 pixels) were extracted from the Train set (good: 28534, poor: 28881), and used a pretrained U-net to get the heat map of nuclei segmentation for each tile before training our model. We concatenated the heatmap of nuclei segmentation and the color-normalized RGB tiles in channel level and produce a 4-channel tile. Bags of 4-channel tiles were then dumped into a feature extractor of the MobileNet V2 model. We used generalized mean with sign as the aggregation function since it is able to keep the extremes while taking the average into account. Output of the aggregation function, which represents score of the pathological image, was activated by a sigmoid function and compared with a given threshold 0.4457, where 0.4457 is also a hyper parameter, and finally classified into certain a class. Pipeline of MobileNet V2 HCC classification (MobileNet V2\_HCC\_Class) is shown in Fig. \ref{pip}.

\subsection{Model generalizes to the LT for HCC dataset}

\begin{figure*}[!htp]
	\centerline{\includegraphics[width=5 in]{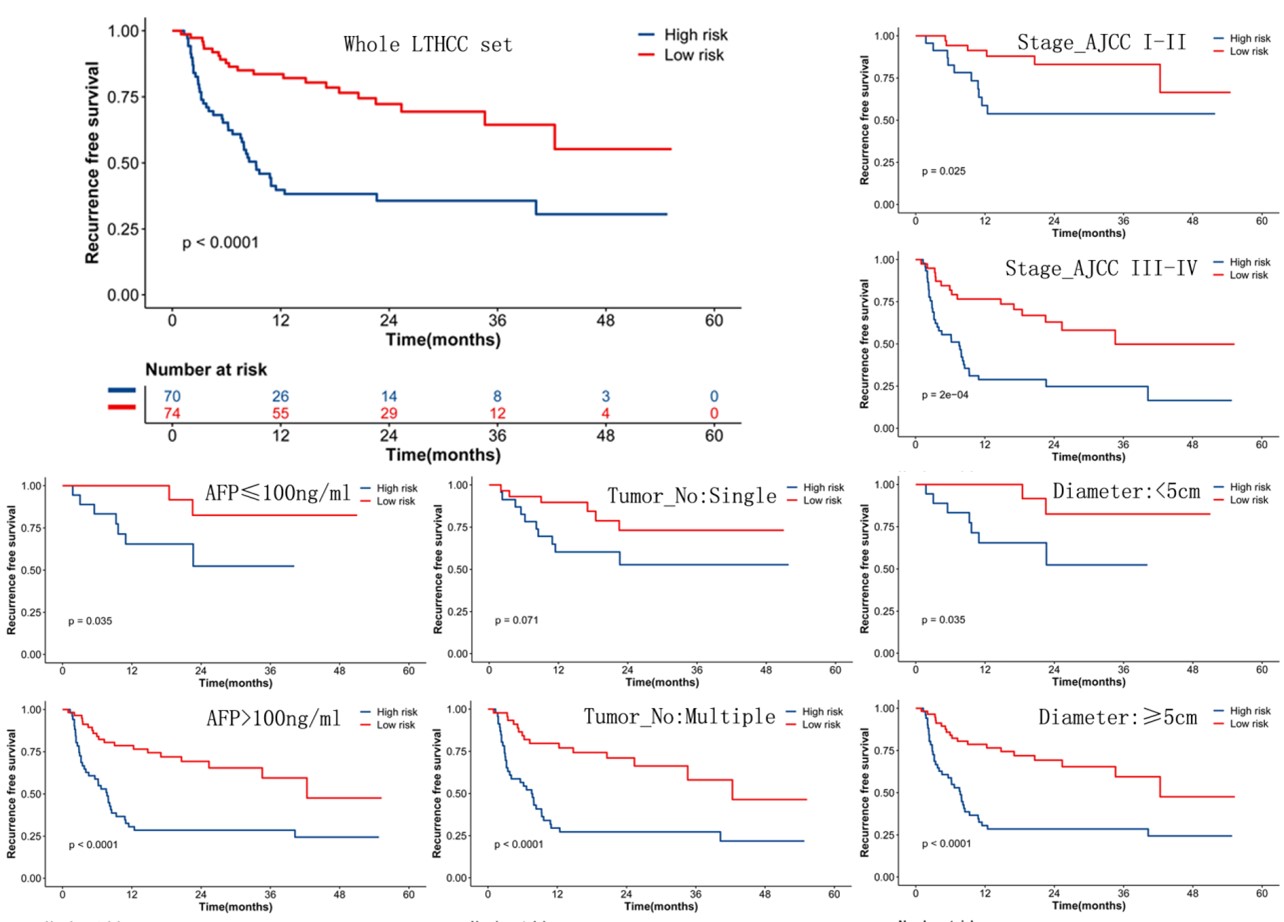}}
	\caption{Prognostic value of MobileNetV2\_HCC\_Class in the whole LT set and after stratification for common prognostic variables. The MobileNetV2\_HCC\_Class categorizes patients into low-risk and high-risk subgroups. The prognostic value of MobileNetV2\_HCC\_Class was conserved, even after stratification according to common clinical and pathological variables. AFP: alpha fetoprotein. Tumor\_No: tumor number. Diameter: total tumor diameter.}
	\label{prog}
\end{figure*}

The output of our neural networks can stratify patients into low and high-risk subgroups. In the LT set, 144 patients with complete follow up data were included, and 65 patients relapsed during follow up. The following variables were available and included in the analysis: age at diagnosis, gender, serum alpha-fetoprotein (AFP), Child-Pugh score, Model for end-stage liver disease (MELD), tumor size, tumor number, grade, tumor stage according to the American Joint Committee on Cancer (Stage AJCC). Univariable analyses indicated that AFP, tumor size, grade, tumor number, and Stage AJCC were associated with shorter RFS (Table \ref{S1}). Tiles from the tissue array of these patients were extracted and processed by our model. The MobileNetV2\_HCC\_Class was a strong predictor of RFS in the whole LT set, even after stratification for other common prognostic features (Stage\_AJCC, AFP, tumor number, and tumor size) (Fig. \ref{prog}).

\begin{table}[h]
	\centering
	\begin{tabular}{|l|l|l|l|}
		\hline
		\textbf{varible}                                         & \textbf{HR} & \textbf{z}   & \textbf{pvalue} \\ \hline
		\textbf{age (year)}                                      & 0.988096274 & -0.923594042 & 0.355697717     \\ \hline
		\textbf{gender (female vs male)}                         & 0.563936442 & -1.108281221 & 0.267740383     \\ \hline
		\textbf{AFP (ng/ml)}                                     & 1.000017536 & 3.134090145  & 0.001723879     \\ \hline
		\textbf{Child-Pugh score}                                & 1.020765101 & 0.352137228  & 0.724735351     \\ \hline
		\textbf{MELD}                                            & 0.994072544 & -0.569040593 & 0.569328593     \\ \hline
		\textbf{MobileNetV2\_HCC\_Class (low risk vs high risk)} & 0.319479798 & -4.28269507  & 1.85E-05        \\ \hline
		\textbf{Tumor\_num (single vs multiple)}                 & 0.459249629 & -2.697250719 & 0.006991461     \\ \hline
		\textbf{Total\_dia ($\geq$cm vs $\textless$5cm)}            & 2.501623191 & 2.546786366  & 0.010871997     \\ \hline
		\textbf{Stage\_TNM (StageI/II vs StageIII/IV)}           & 0.355468852 & -3.582332087 & 0.000340541     \\ \hline
		\textbf{Grade (Well/moderate vs poor)}                   & 0.516786244 & -1.982646675 & 0.04740691      \\ \hline
	\end{tabular}
	\caption{Univariable recurrence free survival analyses in LT set. CHILD, Child-Pugh score; MELD, Model for end-stage liver disease; AFP, serum alpha fetoprotein; Stage\_AJCC, the American Joint Committee on Cancer.}
	\label{S1}
\end{table}

\begin{figure*}[t]
	\centerline{\includegraphics[width=4 in]{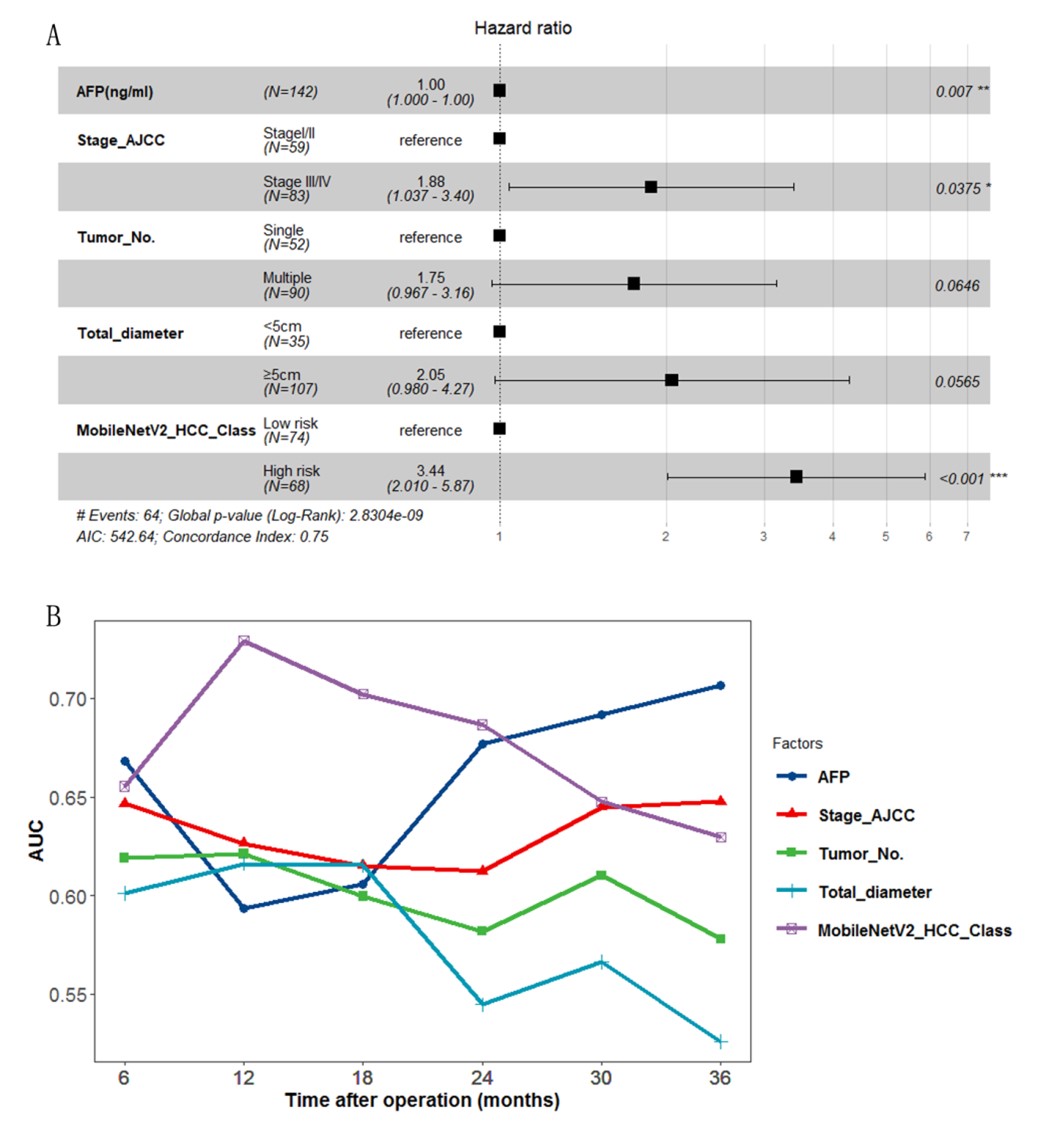}}
	\caption{The performance of different risk factors for tumor recurrence after LT. A. Multivariate analyses. B. The time-dependent area under the receiver operating characteristic curve (AUC) value for different criteria according to tumor recurrence.}
	\label{risk}
\end{figure*}

Multivariate analyses showed that MobileNetV2\_HCC\_Class was of independent prognostic value (HR=3.44 (2.01-5.87), p$\textless$0.001) adjusting for established prognostic markers significant in univariable analyses: Stage\_AJCC, AFP, tumor number, and tumor size (Fig. \ref{risk}A). The time-dependent AUC curves are depicted in Fig. \ref{risk}B. During the entire course of 3-year follow-up, the MobileNetV2\_HCC\_Class maintained relatively higher AUC values than the other factors in the first two years after LT.

\begin{figure*}[h]
	\centerline{\includegraphics[width=5 in]{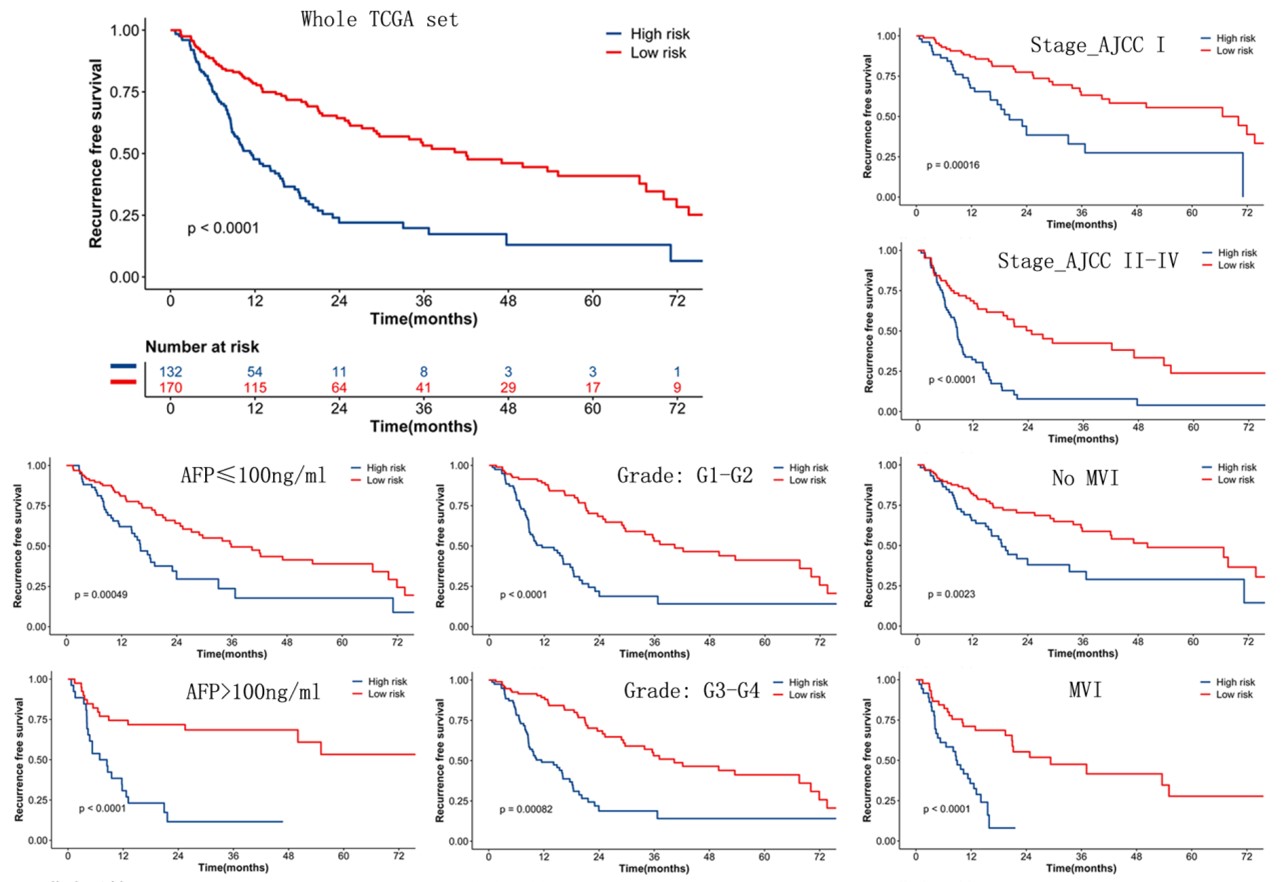}}
	\caption{Prognostic of MobileNetV2\_HCC\_Class in the whole TCGA set and after stratification for common baseline variables. The MobileNetV2\_HCC\_Class predict RFS after stratification for common baseline variables.}
	\label{TCGA}
\end{figure*}

\begin{figure}[h]
	\centerline{\includegraphics[width=4 in]{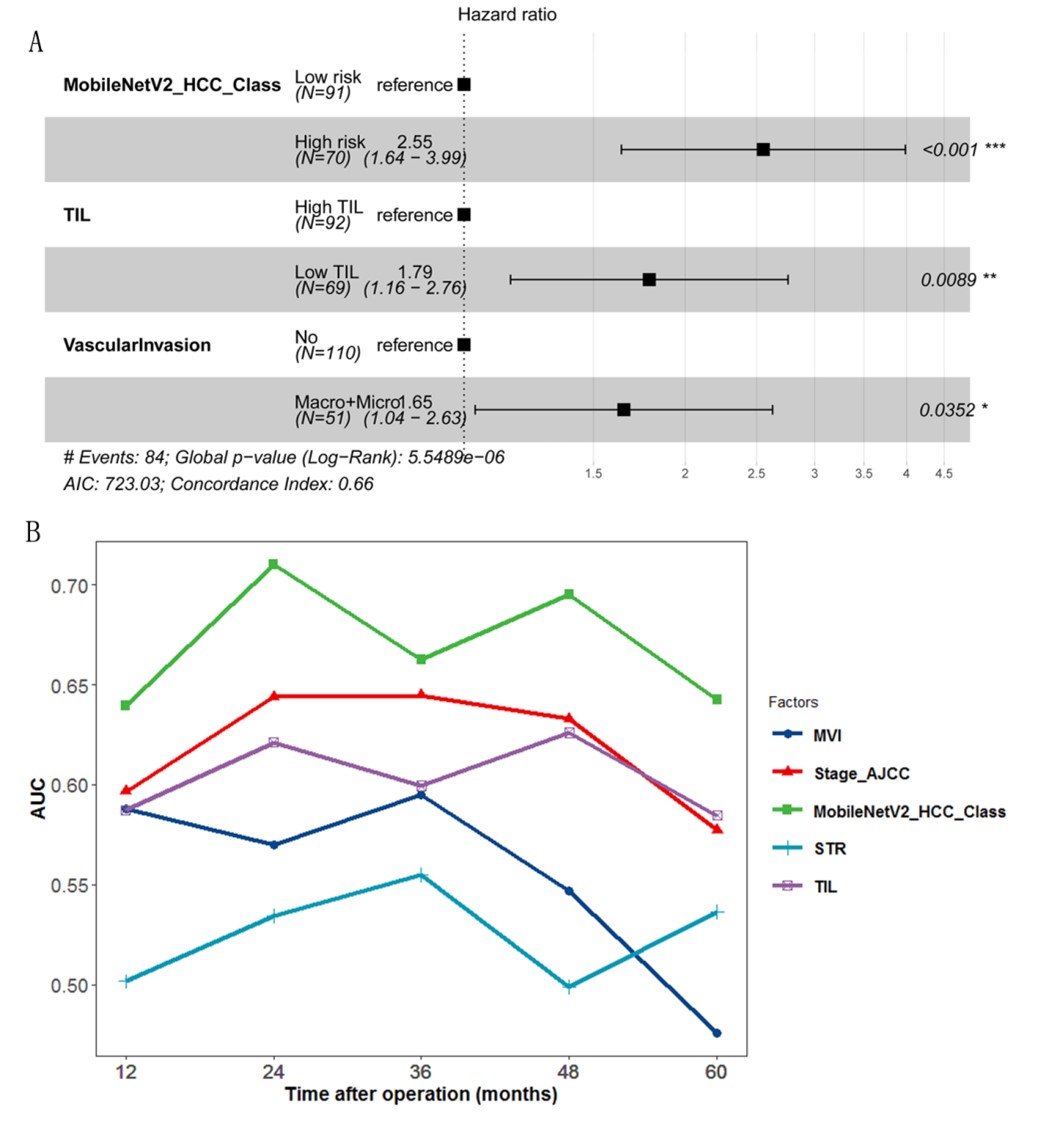}}
	\caption{The performance of different risk factors for tumor recurrence after resection. A. Multivariate analyses. B. The time-dependent area under the receiver operating characteristic curve (AUC) value for different criteria according to tumor recurrence.}
	\label{TCGA_risk}
\end{figure}

\subsection{Model generalizes to the TCGA dataset}
The robustness of the model on an independent series from the TCGA was assessed. In total, 302 patients met the inclusion criteria, and 165 patients with recurrence were recorded. The slides were collected from different centers. The following variables were available and included in the analysis: age at diagnosis, age, gender, AFP, vascular invasion, stroma tumor ratio (STR), tumor-infiltrating lymphocyte (TIL), grade, Stage\_AJCC. The clinical, biological, and pathological features associated with shorter survival were AJCC stage in univariable analyses (Table. \ref{TCGA-tab}). Tiles from WSIs of the 302 patients were extracted and processed by our model. In the TCGA set, MobileNetV2\_HCC\_Class predicted RFS even after stratification for other significant prognostic features (such as Stage\_AJCC, AFP, grade, and vascular invasion) (Fig. \ref{TCGA}).

\begin{table}[]
	\centering
	\begin{tabular}{|l|l|l|l|}
		\hline
		Variables                     & HR       & z        & P value  \\ \hline
		Age (years)                   & 0.995675 & -0.69645 & 0.48615  \\ \hline
		Gender (female vs male)       & 0.859213 & -0.91769 & 0.358779 \\ \hline
		AFP (positive vs negative)    & 1.379967 & 1.770971 & 0.076566 \\ \hline
		vascular invasion (no vs yes) & 0.543518 & -3.32539 & 0.000883 \\ \hline
		TIL (low vs high)             & 1.646678 & 2.72641  & 0.006403 \\ \hline
		STR (low vs high)             & 0.69751  & -2.29362 & 0.021812 \\ \hline
		Stage\_AJCC (II vs I)         & 1.937217 & 3.171588 & 0.001516 \\ \hline
		Stage\_AJCC (III/IV vs I)     & 3.078974 & 5.909885 & 3.42E-09 \\ \hline
		MobileNetV2\_HCC\_Class       & 2.724386 & 6.169901 & 6.83E-10 \\ \hline
	\end{tabular}
	\caption{Univariable recurrence free survival analyses in TCGA set. AFP, serum alpha fetoprotein; STR, stroma tumor ratio; TIL, tumor infiltrating lymphocyte; Stage\_AJCC, the American Joint Committee on Cancer.}
	\label{TCGA-tab}
\end{table}

The classifier remained strong in multivariable analysis (HR=2.55 (1.64-3.99), p$\textless$0.001) adjusting for established prognostic markers significant in univariable analyses: Stage\_AJCC, AFP, grade, and vascular invasion (Fig. \ref{TCGA_risk}A). These observations demonstrate that the model captures complex patterns non-redundant with baseline variables known to affect survival in patients with HCC. The time-dependent AUC curves are depicted in Fig. \ref{TCGA_risk}B. During the entire course of 6-year follow-up, the MobileNetV2\_HCC\_Class maintained relatively higher AUC values than the other factors after HCC resection.

\subsection{Histological analysis of tiles}
\begin{figure*}[h]
	\centerline{\includegraphics[width=4 in]{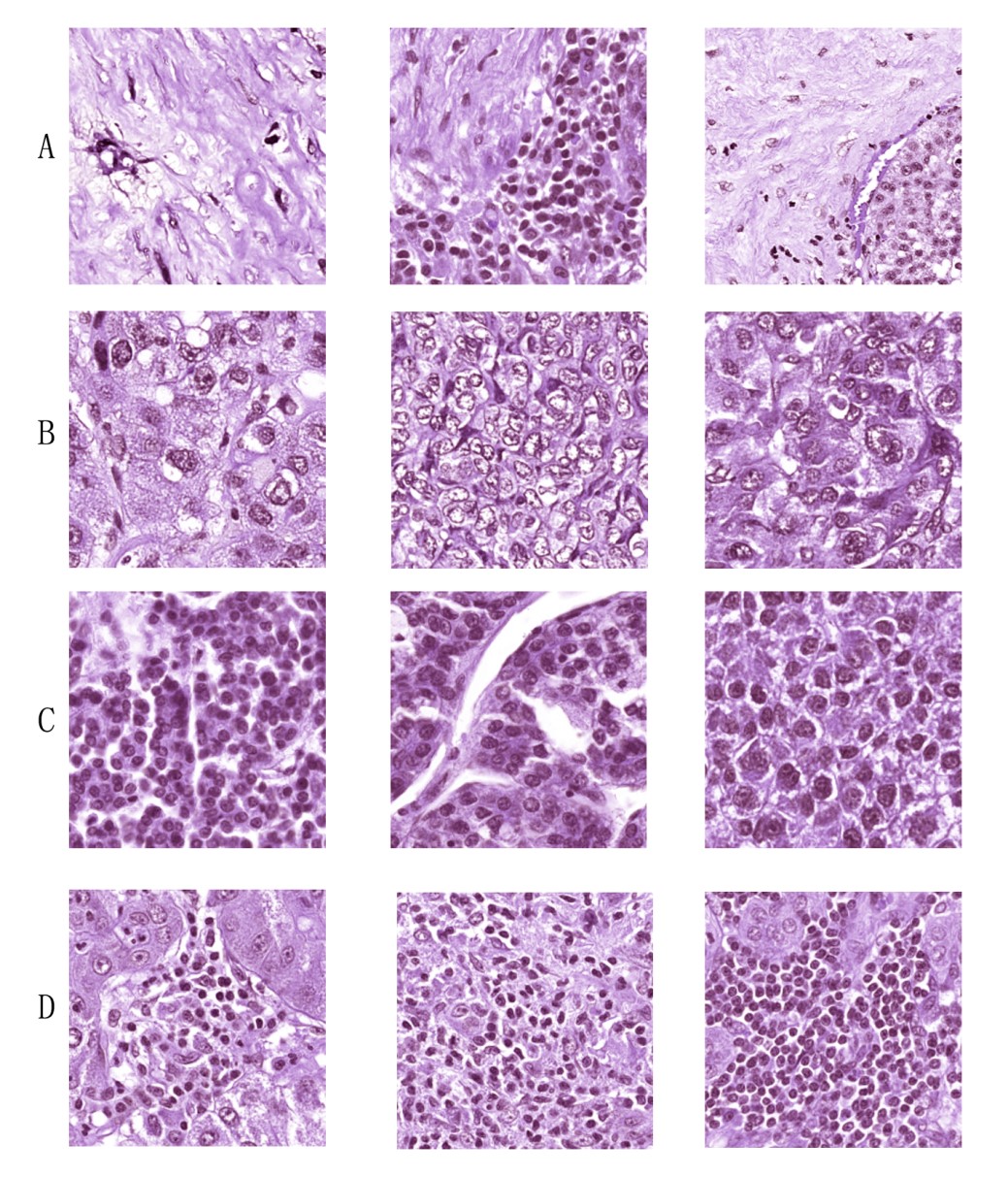}}
	\caption{ Typical tiles of classified as low or high risk by MobileNetV2\_HCC\_Class. 400 most predictive tiles were analyzed. The features predictive of a high risk of recurrence included stroma (A), cellular atypia (B), and nuclear hyperchomasia (C). The feature predictive of a low risk of recurrence included the presence of immune cells (D).}
	\label{his}
\end{figure*}

The MobileNetV2\_HCC\_Class can retrieve the most predictive tiles from the thousands of tiles processed. The main histological features of recurrence were investigated by extracting the 400 most predictive tiles (high risk of recurrence: 200, low risk of recurrence: 200) from 302 patients of the TCGA with MobileNetV2\_HCC\_Class. Four histological features were recorded in tumoral areas. Presence of stroma, high degree of cytological atypia, and nuclear hyperchomasia were associated with high risk (p=0.0003, p=0.0010, p=0.0012, respectively), while immune cell infiltration was associated with low risk (p=0.0019) (Fig. \ref{his}, Table. \ref{tab_his}). These results show that our deep learning model can also detect the known histological patterns associated with recurrence in HCC patients.

\begin{table}[h]
	\centering
	\begin{tabular}{|l|l|l|l|}
		\hline
		Features                          & High risk & Low risk & P value \\ \hline
		Presence of stroma                & 25/200    & 4/200    & 0.0003  \\ \hline
		High degree of cytological atypia & 36/200    & 11/200   & 0.001   \\ \hline
		Nuclear hyperchomasia             & 28/200    & 7/200    & 0.0012  \\ \hline
		Immune cell infiltration          & 3/200     & 19/200   & 0.0019  \\ \hline
	\end{tabular}
	\caption{Histological features in tiles associated with recurrence.}
	\label{tab_his}
\end{table}

\section{Discussion}
Building on recent developments in deep learning, we have developed MobileNetV2\_HCC\_Class for automatic prediction of prognosis of post-operation HCC patients, which directly analyses standard histological sections stained with H\&E. Our algorithms predict prognosis more accurately than classical clinical, biological and pathological features.

Deep learning-driven approaches for processing medical images have already been shown to standardize the diagnosis of cancer and improve patient stratification\cite{bejnordi2017diagnostic}\cite{niazi2019digital}. In a recent pioneering study, a deep learning-based model could detect and classify lung cancer with similar accuracy as pathologists\cite{coudray2018classification}. Previous studies suggest that deep learning could be used to develop markers that potentially use basic morphology to predict the outcome of patients with cancer\cite{mobadersany2018predicting}\cite{bychkov2018deep}. A deep learning-based model by Coudray and coworkers could predict six of the most frequent genetic alterations directly from the slides\cite{coudray2018classification}. In gastrointestinal cancer, deep learning can enable estimation of  microsatellite instability directly from histology images\cite{kather2019deep}. Kather and co-workers reported that CNN extracts the tumor components and predicts patient survival directly from histology images\cite{kather2019predicting}. Saillard et al. used CNN to predict the survival in HCC which extracted the features from the images by a pretrained CNNs and the network then selected 25 tiles with the highest and lowest scores for the prediction of patient survival\cite{saillard2020predicting}. In our study, a different method was used to develop MobileNetV2\_HCC\_Class to improve the prediction of prognosis in HCC treated by surgical resection and LT. The scientific or innovation points of our method are: random titles of each patient were used, like Skrede\cite{skrede2020deep}, trained MobileNet V2 by MIL which allow training on tile collections labelled with the label of its whole-slide image; and most importantly, nuclear architectural information was used in building model, which was useful in cancer grading and predicting patient out-comes\cite{ji2019nuclear}. Genetic instability could be displayed by diversify of nuclear shape and texture, playing important role in metastasis and proliferation that potentially results in cancer recurrence. MobileNetV2\_HCC\_Class was a strong predictor of RFS in the HCC patient treated with resection or LT, and generalized in the TCGA set across different centers.

Molecular and/or genetic features can predict the survival of patients with HCC\cite{zucman2015genetic}\cite{pan2020biomarkers}. Chaudhary et al. used deep-learning approaches on RNA sequencing and methylation data from the TCGA database to explicitly predict HCC survival from multiple patient cohorts. The high-throughput gene expression profiling/sequencing technologies in clinical practice are currently hampered by their cost and reproducibility issues. Our model only requires histological slides which are routinely available in surgical treatment centers. Additionally, the histological slide processing and computing time should also be short enough to avoid delaying therapeutic decisions. This approach would thus facilitate the implementation of risk stratification systems in clinical practice. 

In summary, a clinically useful prognostic model was developed using deep learning allied to histological slides. The model has been extensively evaluated in independent patient populations with different treatment and gives consistently excellent results across the classical clinical, biological, and pathological features. The proposed CNN-based approach may lead to improvements in the assessment of patient prognosis, and help guide clinicians in their decision-making about the use of adjuvant therapy.

\section*{Acknowledgments}
The study was supported by Key Program of National Natural Science Foundation of China (No. 81930016)), National Natural Science Foundation of China (No. 81802889). The results in this study are in part based upon data from the TCGA.

\bibliographystyle{./elsarticle-num.bst}
\bibliography{references}  






\end{document}